\documentclass{article}
\usepackage{spconf,amsmath,graphicx,hyperref}
\usepackage{threeparttable}
\usepackage{booktabs}

\title{CAF-Mamba: Mamba-Based Cross-Modal Adaptive Attention Fusion for Multimodal Depression Detection}
\name {Bowen Zhou, Marc-André Fiedler, and Ayoub Al-Hamadi\thanks{This work was supported by the German Research Foundation (DFG) under Project AL 638/13-3 (Grant No. 468478819) and by the European Regional Development Fund (ERDF) within the Project ORAKEL (Grant No. ZS/2023/12/182322).}
\thanks{© 2026 IEEE. Personal use of this material is permitted.
Permission from IEEE must be obtained for all other uses, in any
current or future media, including reprinting/republishing this
material for advertising or promotional purposes, creating new
collective works, for resale or redistribution to servers or lists,
or reuse of any copyrighted component of this work in other works.}
\thanks{This paper has been accepted for publication in the proceedings of
2026 IEEE ICASSP Conference.}}

\address{Neuro-Information Technology Group (NIT)\\
        IIKT, Otto von Guericke University Magdeburg\\
	    Magdeburg, Germany\\
        bowen.zhou@ovgu.de}

\begin{document}
\topmargin=0mm
\ninept
\maketitle
\begin{abstract}
Depression is a prevalent mental health disorder that severely impairs daily functioning and quality of life. While recent deep learning approaches for depression detection have shown promise, most rely on limited feature types, overlook explicit cross-modal interactions, and employ simple concatenation or static weighting for fusion. To overcome these limitations, we propose CAF-Mamba, a novel Mamba-based cross-modal adaptive attention fusion framework. CAF-Mamba not only captures cross-modal interactions explicitly and implicitly, but also dynamically adjusts modality contributions through a modality-wise attention mechanism, enabling more effective multimodal fusion. Experiments on two in-the-wild benchmark datasets, LMVD and D-Vlog, demonstrate that CAF-Mamba consistently outperforms existing methods and achieves state-of-the-art performance. Our code is available at https://github.com/zbw-zhou/CAF-Mamba.
\end{abstract}
\begin{keywords}
Adaptive attention, cross-modal, depression detection, mamba, multimodal fusion
\end{keywords}
\section{Introduction}
\label{sec:intro}

Depression is a widespread mental health disorder and is expected to become the most common mental disorder by 2030~\cite{MALHI20182299}. Early warning and accurate detection are therefore essential for effective intervention and treatment. In recent years, machine learning and deep learning techniques have drawn increasing research attention in the healthcare and medical fields~\cite{CHAKRABORTY,Fiedler2023,Fiedler}, and numerous studies have been proposed for multimodal depression detection. TAMFN~\cite{TAMFN} employs a time-aware attention mechanism with TCN and LSTM for multimodal depression detection. Transformers have been widely adopted to model cross-modal interactions and fuse multimodal data~\cite{lmvd,dvlog,STST}. Recent advances, such as Mamba, have demonstrated robustness in modeling long sequences~\cite{mamba,ViM,stcmmamba,cmmamba}. Ye et al.~\cite{depmamba} proposed a bidirectional variant with shared parameters to capture intermodal dependencies for depression detection, while separate Mamba encoders have been employed to derive cross-modal fusion in MDDMamba~\cite{MDDMamba}. Furthermore,  compared with depression datasets recorded in controlled laboratory settings through interviews, recent studies utilize datasets collected from social media platforms, which may better cover usual behaviors of depressed individuals in the wild~\cite{GADZAMA,Tahir}.

Despite significant progress, several limitations remain in existing multimodal depression detection approaches. First, many existing methods rely primarily on audio features and facial landmarks as visual features, while other informative features remain underexplored. For example, facial Action Units (AUs) represent movements of facial muscles~\cite{ekman1978facs}, whereas eye-related and head movement features (eye landmarks, gaze, head pose, etc.) may provide valuable behavioral indicators of depression. Second, explicit modeling of cross-modal interactions is often absent, which limits the capacity to exploit complementary relationships among modalities. Third, many conventional fusion strategies employ modalities with uniform static weighting or simple concatenation, without accounting for the fact that the importance and quality of different modalities may differ across samples in real-world scenarios.

\begin{figure*}[htb]

\begin{minipage}[b]{1.0\linewidth}
  \centering
  \centerline{\includegraphics[width=16.8cm]{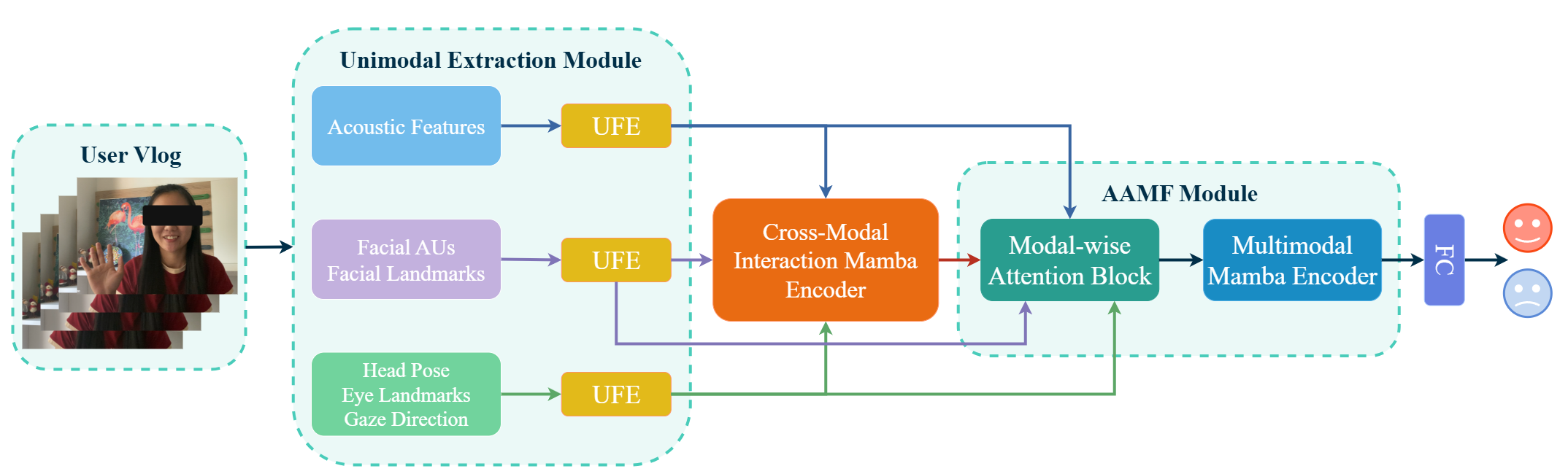}}
  \centerline{(a) The framework of CAF-Mamba}
  \label{fig:framework}
\end{minipage}
\begin{minipage}[b]{0.3\linewidth}
  \raggedright
  \includegraphics[width=3.5cm]{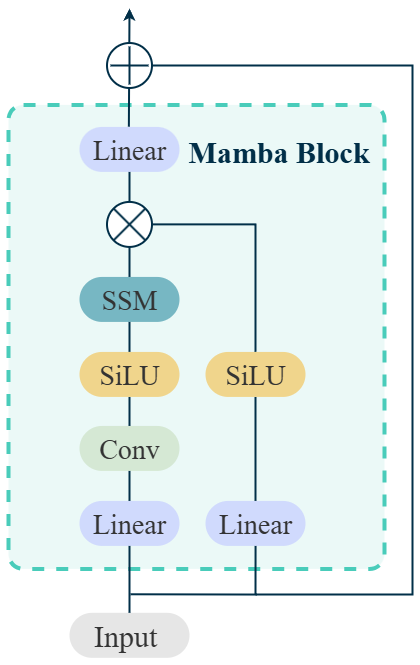} 
  \centerline{(b) ResMamba}\medskip
  \label{fig:resmamba}
\end{minipage}
\hfill
\begin{minipage}[b]{0.68\linewidth}
  \raggedleft
  \includegraphics[width=\linewidth]{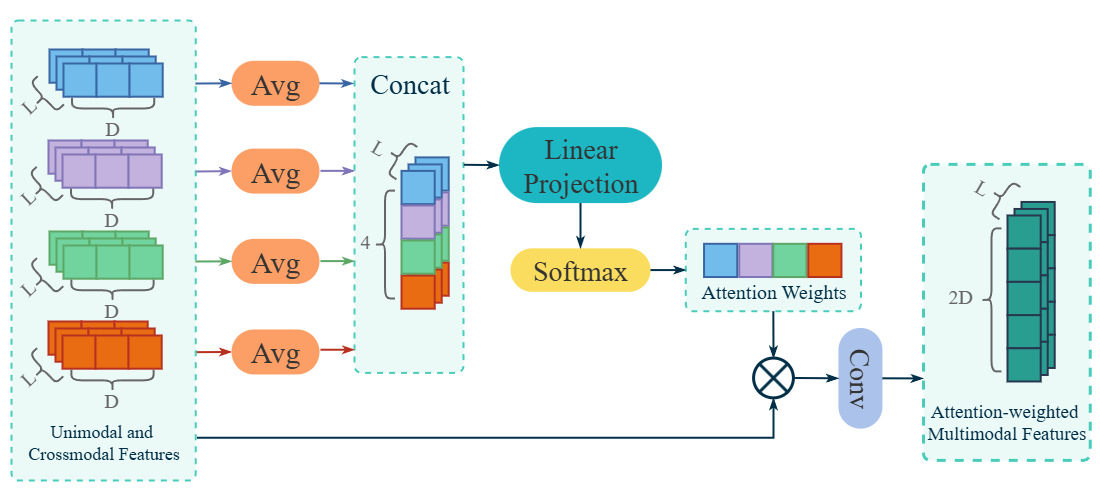}
  \centerline{(c) Modal-wise Attention Block}\medskip
  \label{fig:attention}
\end{minipage}
\vspace{-2mm}
\caption{Overview of CAF-Mamba.}
\label{fig:overview}
\end{figure*}

To address these challenges, we propose a Mamba-based cross-modal adaptive attention fusion (\textbf{CAF-Mamba}) framework for multimodal depression detection. Leveraging the Mamba architecture, we design three key components within CAF-Mamba: a Unimodal Extraction Module (\textbf{UEM}) to extract unimodal representations for each modality, a Cross-modal Interaction Encoder (\textbf{CIME}), which \textbf{explicitly} captures inter-modal dependencies, and an Adaptive Attention Mamba Fusion Module (\textbf{AAMFM}), which \textbf{implicitly} models higher-order correlations and employs a \textbf{modality-wise attention mechanism} to dynamically adjust the importance of each modality, thereby emphasizing relevant information while suppressing noisy or redundant signals. Through hierarchical representation learning, the overall pipeline evolves from unimodal feature extraction to intermodal interaction and finally to adaptive attention fusion, enabling the multimodal representation to capture more comprehensive and complementary cues for depression detection. We validate the framework on two in-the-wild benchmark datasets, where CAF-Mamba consistently outperforms existing methods and achieves state-of-the-art (SOTA) performance. Furthermore, the framework can be extended to an arbitrary number of modalities, ensuring broad applicability. 


\section{Methodology}
The overview of the CAF-Mamba framework is illustrated in Fig. \ref{fig:overview}, and detailed descriptions of each component are provided in the following sections.

\subsection{Preliminaries on Mamba}
The State Space Model (SSM)~\cite{s4} originates from linear time-invariant systems, where input sequences are mapped to outputs via intermediate hidden states. To adapt SSM for deep learning, discretized variants such as  Structured State Space Models (S4) and Mamba have been developed. These models reformulate the continuous system into linear recursive and convolutional forms, enabling efficient sequence modeling. Notably, Mamba enhances SSM with a selective mechanism that makes its parameters input-dependent, dynamically emphasizing relevant information while ignoring irrelevant information, thereby further improving its ability to capture long-range dependencies and complex correlations.

\subsection{Unimodal Extraction Module}
The UEM comprises three Unimodal Feature Extractors (UFEs), each responsible for modeling temporal dependencies and modality-specific representations. Each UFE contains a 1D CNN to project the input into a unified embedding space, followed by a ResMamba structure, in which a residual connection is integrated into the Mamba block as shown in Fig. \ref{fig:resmamba}(b). This design simultaneously preserves low-level features and high-level abstract representation, while also stabilizing the training of models.
\begin{equation}
X'_a, X'_{lau}, X'_{egh} = \mathrm{UFEs}(X_a,X_{lau},X_{egh}),
\end{equation}
$X_a$, $X_{lau}$ and $X_{egh}$ denote acoustic features, facial landmarks with AUs, and eye-gaze-head features, respectively, each fed into its corresponding UFE.

\subsection{Cross-Modal Interaction Mamba Encoder}
The CIME is designed to explicitly capture interactions among modalities, complementing unimodal representations that preserve modality-specific cues but may miss intermodal correlations. Since unimodal features describe the same subject state, strong correlations often exist, for instance, sadness is often reflected by a low-pitched voice, downturned mouth corners and a downward gaze. CIME takes the extracted unimodal features $X'_a$, $X'_{lau}$, $X'_{egh}$ as inputs and computes an additional representation $X_i$ that encodes cross-modal interactions. It is built upon the ResMamba structure, 
\begin{equation}
X_i = \mathrm{ResMamba}\big(\mathrm{Add}(X'_a, X'_{lau}, X'_{egh})\big),
\end{equation}
where $X_i$ is the intermodal representation.

\subsection{Adaptive Attention Mamba Fusion Module}
The AAMFM is designed to fuse unimodal and intermodal representations dynamically and contextually into a multimodal representation for classification. As illustrated in Fig. \ref{fig:overview}(a), the  AAMFM consists of two key components, the Modal-wise Attention Block (MAB) and the Multimodal Mamba Encoder (MME).

\vspace{0.5em}
\subsubsection{Modal-wise Attention Block}
Unlike conventional fusion methods that apply simple concatenation or static weighting, an adaptive attention mechanism is designed in the MAB to assign dynamic weights to both unimodal and intermodal features. As shown in Fig. \ref{fig:attention}(c), extracted features are first processed with average pooling along the temporal dimension to obtain compact representations, which are then concatenated and passed through a learnable linear projection followed by Softmax to generate attention weights $\alpha$. These weights are subsequently applied to emphasize important modalities while suppressing less relevant ones, and a 1D convolutional layer projects the fused features into the intermediate attention-weighted representation $X'$:
\begin{align}
\alpha &= \sigma\Big( W \big[ \mathrm{A}(X'_a) \,\|\, \mathrm{A}(X'_{lau}) \,\|\, \mathrm{A}(X'_{egh}) \,\|\, \mathrm{A}(X_i) \big] \Big), \\
X' &= \mathrm{Conv}\big( \alpha_1 X'_a \,\|\, \alpha_2 X'_{lau} \,\|\, \alpha_3 X'_{egh} \,\|\, \alpha_4 X_i \big),
\end{align}
where $W$ denotes a learnable linear projection, $A$ denotes average pooling, and $||$ means concatenation.

\vspace{0.5em}

\subsubsection{Multimodal Mamba Encoder}
MME is employed to model global long-range temporal dependencies of the attention-weighted multimodal features and to capture high-order cross-modal correlations implicitly. Similar to UFE, the MME is built upon the ResMamba structure. The input $X'$ is processed by the MME to generate the final fused multimodal representation $M$. Finally, a linear layer is employed to perform depression detection. 

\section{Experiments}

\begin{table}[!t]
\centering
\caption{Performance Comparison of Baseline and Proposed Models Using Multimodal Features on LMVD (\%)}
\label{tab:multimodal}
\begin{threeparttable}
\begin{tabular}{c|cccc}
\toprule
\textbf{Method} & Accuracy & Precision & Recall & F1\\ 
\midrule
KNN~\cite{KNN}                              & 58.34 & 59.75 & 58.34 & 56.87 \\
SVM~\cite{SVM}                              & 64.66 & 65.77 & 64.66 & 64.06 \\
LR~\cite{LR}                                & 64.88 & 65.19 & 64.88 & 64.73 \\
RF~\cite{RF}                                & 69.23 & 69.34 & 69.23 & 69.17 \\
Xception\cite{xception}                     & 71.38 & 71.94 & 71.38 & 71.19 \\  
SEResnet\cite{seresnet}                     & 72.54 & 73.13 & 72.54 & 72.36 \\ 
BiLSTM~\cite{BLSTM}                         & 72.59 & 73.02 & 72.59 & 72.47 \\ 
ViT~\cite{ViT}                              & 73.03 & 73.52 & 73.03 & 72.90 \\
MDDformer~\cite{lmvd}                       & 76.88 & 77.02 & 76.88 & 76.85 \\
\midrule
\textbf{CAF-Mamba}     & \textbf{78.69} & \textbf{78.26} & \textbf{79.12} & \textbf{78.69}\\
\bottomrule
\end{tabular}
\end{threeparttable}
\end{table}

\begin{table*}[htbp]
\centering
\caption{Performance Comparison of Baseline and Proposed Models Using Bimodal Features on LMVD and D-Vlog (\%)}
\label{tab:bimodal}
\begin{tabular}{c|ccccc|ccccc}
\toprule
& \multicolumn{5}{c|}{\textbf{LMVD}} & \multicolumn{5}{c}{\textbf{D-Vlog}} \\
\textbf{Method} & Accuracy & Precision & Recall & F1 & Average  
& Accuracy & Precision & Recall & F1 & Average \\
\midrule
KNN~\cite{KNN}                   & 56.83 & 57.43 & 50.92 & 53.92 & 54.77 & -     & 57.86 & 59.43 & 54.25 & 57.18 \\
BLSTM~\cite{BLSTM}               & 66.85 & 65.81 & 70.33 & 67.83 & 67.71 & -     & 60.81 & 61.79 & 59.70 & 60.77 \\
TFN~\cite{TFN}                   & 63.93 & 64.08 & 62.64 & 63.34 & 63.50 & -     & 61.39 & 62.26 & 61.00 & 61.55 \\
TAMFN\cite{TAMFN}                & 70.49 & 71.15 & 68.86 & 69.84 & 70.09 & -     & 66.02 & 66.50 & 65.82 & 66.11 \\
DepDetector~\cite{dvlog}         & 61.93 & 60.36 & 72.16 & 65.08 & 64.88 & -     & 65.40 & 65.57 & 63.50 & 64.82 \\
STST\cite{STST}                  & 67.76 & 69.20 & 64.01 & 66.23 & 66.80 & 70.70 & 72.50 & 77.67 & 75.00 & 73.97 \\
DepMamba\cite{depmamba}          & 72.13 & 70.18 & 76.56 & 73.20 & 73.02 & 68.87 & 68.19 & \textbf{86.99} & 76.44 & 75.12 \\
\midrule
\textbf{CAF-Mamba}  & \textbf{74.32} & \textbf{72.92} & \textbf{76.92} & \textbf{74.87} & \textbf{74.76} 
                    & \textbf{72.17} & \textbf{73.88} & 80.49 & \textbf{77.04} & \textbf{75.90} \\
\bottomrule
\end{tabular}
\end{table*}

\begin{table*}[htbp]
\centering
\caption{Efficiency Comparisons between Transformer-based DepDetector and CAF-Mamba}
\label{tab:efficiency}
\begin{tabular}{c|c|cccccccccc}
\toprule
& \multicolumn{1}{c|}{\textbf{Parameters}} & \multicolumn{10}{c}{\textbf{Inference Speed at Different Sequence Lengths (ms)}} \\
\textbf{Method} & (M) & 1000 & 2000 & 3000 & 4000 & 5000 & 6000 & 7000 & 8000 & 9000 & 10000 \\
\midrule
DepDetector                   & 1.06 & 3.49 & 3.58 & 3.55 & 3.96 & 4.84 & 5.81 & 6.74 & 7.86 & 12.67 & 14.16 \\
CAF-Mamba                     & 0.57 & 1.42 & 1.43 & 1.46 & 1.67 & 2.29 & 2.47 & 2.94 & 3.20 & 3.99 & 4.32\\
\bottomrule
\end{tabular}
\end{table*}

\begin{table}[!t]
\centering
\caption{Ablation Study on Modality Combinations and the Effectiveness of CIME and AAMFM (\%)}
\label{tab:ablation}
\begin{threeparttable}
\begin{tabular}{c|cccc}
\toprule
\textbf{Method} & Accuracy & Precision & Recall & F1 \\
\midrule
LAU + EGH                 & 72.13 & 69.61 & 78.02 & 73.58 \\
A + LM                    & 74.32 & 72.92 & 76.92 & 74.87 \\
A + LAU                   & 75.41 & 73.96 & 78.02 & 75.94 \\
A + EGH                   & 75.96 & 74.23 & 79.12 & 76.60 \\
w/o CIME                  & 73.22 & 71.43 & 76.92 & 74.07 \\
w/o AAMFM                 & 74.32 & 72.45 & 78.02 & 75.13 \\
\midrule
CAF-Mamba                & 78.69 & 78.26 & 79.12 & 78.69 \\
\bottomrule
\end{tabular}
\begin{tablenotes}
\small
\item w/o = without; A = acoustic features; LM = facial landmarks; LAU = facial landmarks + AUs; EGH = eye-gaze-head features.
\end{tablenotes}
\end{threeparttable}
\end{table}

\subsection{Experimental Setup}
\subsubsection{Dataset}
To evaluate the effectiveness of the proposed method, we conduct experiments on two public datasets: the Large-Scale Multimodal Vlog Dataset (LMVD)~\cite{lmvd} and the Depression Vlog Dataset (D-Vlog)~\cite{dvlog}. Both datasets are collected from social media and provide only pre-extracted modality-specific features due to privacy constraints. D-Vlog comprises 961 vlogs from 816 individuals, totaling about 160 hours of video, and provides 68 facial landmarks extracted using dlib~\cite{dlib} as visual features and 25 low-level descriptors (LLDs) extracted using OpenSmile~\cite{opensmile} as acoustic features. LMVD contains 1,823 vlogs from 1,475 participants, with a total duration of about 214 hours. In addition to facial landmarks, LMVD provides AUs, eye landmarks, gaze direction, and head pose as visual features, as well as 128-dimensional audio embeddings extracted using a pre-trained VGGish model. We use LMVD for multimodal experiments and both datasets for bimodal experiments. LMVD is divided into training, validation, and test sets in an 8:1:1 ratio, and D-Vlog is divided in a 7:1:2 ratio. 

\subsubsection{Implementation Details}
Experiments have been conducted on an NVIDIA GeForce RTX 4090 with 24 GB of memory using PyTorch. The model was optimized with Adam using an initial learning rate of 0.0001 and a ReduceLROnPlateau scheduler using a reduction factor of 0.6 for 80 epochs with a batch size of 16. Each of the UFE, CIME, and MME contains a single ResMamba block, in which Mamba is configured with a dimensionality of 256, and the model was trained using Binary Cross-Entropy loss. Model performance was evaluated using accuracy, precision, recall, and F1 score. 

\subsection{Comparison Results}

\subsubsection{Multimodal Features}
To evaluate the performance of CAF-Mamba, we compare it with several baseline methods using multimodal features from the LMVD dataset. The baselines include traditional machine learning approaches, as well as deep learning architectures such as Xception~\cite{xception}, SEResNet~\cite{seresnet} and ViT~\cite{ViT}. As shown in Table \ref{tab:multimodal}, CAF-Mamba achieves the best overall performance, with accuracy of 78.69 \%, precision of 78.26 \%, recall of 79.12 \% and F1 score of 78.69 \%. Notably, compared to MDDformer, which is a Transformer-based model proposed in the LMVD paper, CAF-Mamba yields improvements of 1.81 \% in accuracy, 1.24 \% in precision, 2.24 \% in recall and 1.84 \% in F1 score. These results demonstrate that the CAF-Mamba has the strong ability to capture complementary and comprehensive representations across modalities and fuse multimodal features effectively.

\subsubsection{Bimodal Features}
To further evaluate the performance of CAF-Mamba, the model is trained with the same architecture using only acoustic features and facial landmarks, which are commonly used in recent depression detection studies, and is evaluated on both the D-Vlog and LMVD datasets. As shown in Table \ref{tab:bimodal}, CAF-Mamba achieves superior performance on both datasets. On the D-Vlog dataset, compared to the Mamba-based model DepMamba, our method achieves higher precision with an improved F1 score, which captures a better balance between precision and recall. On the LMVD dataset, CAF-Mamba consistently achieves SOTA results across all metrics.

\subsection{Ablation study}

\subsubsection{Different Modality Combination}
To analyze the impact of different modality combinations, we conduct a series of ablation experiments trained with only two modalities. As shown in Table \ref{tab:ablation}, for bimodal depression detection, the combination of audio and eye-gaze-head features achieves the best performance across all metrics. Compared with other modalities, audio provides more important cues of depression detection, as excluding acoustic features results in a substantial drop in precision. Furthermore, compared with using only facial landmarks as visual features, the inclusion of AUs leads to improved detection performance. In addition, integrating multimodal data leads to a significant performance improvement with a 4.03\% increase in precision over the best bimodal fusion result. These findings confirm that using multimodal information in fusion provides more comprehensive cues for depression detection.

\subsubsection{Effectiveness of CIME and AAMFM}
To further assess the contribution of the explicit cross-modal interaction and the adaptive attention Mamba fusion mechanism, we conduct ablation experiments by systematically removing CIME and AAMFM. As shown in Table \ref{tab:ablation}, eliminating either component results in a substantial performance drop across all metrics. Notably, removing the AAMFM module and applying simple concatenation results in a decrease of 5.81\% in precision and the absence of the CIME results in a greater decline of 6.83\%. These findings clearly demonstrate that the explicit cross-modal interaction is critical for capturing complementary cross-modal representations,  and the Mamba-based adaptive attention mechanism plays a crucial role in effectively fusing multimodal data.

\subsubsection{Efficiency Analysis}
To evaluate the computational efficiency of our framework, we conducted a comparison between CAF-Mamba and the Transformer-based DepDetector, as reported in Table \ref{tab:efficiency}. To obtain the average inference time per run, each model was tested by repeating the inference process 100 times in a single epoch, and the total time was divided by the number of repetitions. The results indicate that our CAF-Mamba model, with approximately half the number of parameters (0.57M), consistently outperforms the DepDetector in terms of inference speed. More importantly, as the sequence length increases, CAF-Mamba exhibits a near-linear increase in inference time, highlighting the advantage of its Mamba-based architecture. In contrast, the DepDetector's inference time shows a more pronounced growth, reflecting the quadratic complexity inherent in traditional Transformer models. This indicates that CAF-Mamba maintains more stable inference-time scaling as input size increases. These results demonstrate that CAF-Mamba offers superior efficiency and scalability, making it particularly suitable for processing long-form data in real-world applications.

\section{Conclusion}
\label{sec:typestyle}
In this paper, we proposed CAF-Mamba, a novel framework for multimodal depression detection that explicitly and implicitly models cross-modal information while dynamically adjusting attention across diverse modalities. Our approach achieves state-of-the-art performance in both multimodal and bimodal fusion experiments. Beyond its accuracy, our framework demonstrates superior efficiency, exhibiting a near-linear increase in inference time with longer sequences. Ablation studies further validate the effectiveness of the framework's key components. For future work, we plan to explore more efficient model architectures and advanced fusion strategies, and to extend evaluations to both laboratory and in-the-wild benchmark datasets in order to further improve the accuracy and generalizability of the framework.

\vfill\pagebreak

\bibliographystyle{IEEEbib}
\bibliography{strings,refs}

\end{document}